\documentclass[10pt,twocolumn,letterpaper]{article}

\usepackage{wacv}
\usepackage{times}
\usepackage{epsfig}
\usepackage{graphicx}
\usepackage{amsmath}
\usepackage{amssymb}
\usepackage{algorithmic}
\usepackage{algorithm}
\usepackage{multirow}
\usepackage{booktabs} 
\usepackage{subfigure}
\usepackage{mathtools}
\usepackage{color}
\usepackage{pdfpages}

\def\wacvPaperID{830} 

\wacvfinalcopy 

\ifwacvfinal
\fi

\ifwacvfinal
\usepackage[breaklinks=true,bookmarks=false]{hyperref}
\else
\usepackage[pagebackref=true,breaklinks=true,colorlinks,bookmarks=false]{hyperref}
\fi

\ifwacvfinal
\else
\fi

\begin{document}

\title{Cross-modal Adversarial Reprogramming}

\author{\text{*}Paarth Neekhara, \text{*}Shehzeen Hussain, Jinglong Du, Shlomo Dubnov, Farinaz Koushanfar, Julian McAuley\\
University of California San Diego\\
{\tt\footnotesize \{pneekhar,ssh028\}@ucsd.edu} \\
\text{*} Equal contribution\\
}

\maketitle
\thispagestyle{empty}

\begin{abstract}
With the abundance of large-scale deep learning models, it has become possible to repurpose pre-trained networks for new tasks.
Recent works on \emph{adversarial reprogramming} have shown that it is possible to repurpose neural networks for alternate tasks without modifying the network architecture or parameters. However these works only consider original and target tasks within the same data domain.
In this work, we broaden the scope of adversarial reprogramming beyond the data modality of the original task.
We analyze the feasibility of adversarially repurposing image classification neural networks for Natural Language Processing (NLP) and other sequence classification tasks.
We design an efficient adversarial program that maps a sequence of discrete tokens into an image which can be classified to the desired class by an image classification model. 
We demonstrate that by using highly efficient adversarial programs, we can reprogram image classifiers to achieve competitive performance on a variety of text and sequence classification benchmarks without retraining the network.

\end{abstract}

\section{Introduction}

Transfer learning~\cite{Raina07self-taughtlearning} and adversarial reprogramming~\cite{reprogramming} are two closely related techniques used for repurposing well-trained neural network models for new tasks. 
Neural networks when trained on a large dataset for a particular task, learn features that can be useful across multiple related tasks. Transfer learning aims at exploiting this learned representation for adapting a pre-trained neural network for an alternate task. 
Typically, 
the last few layers of a neural network are modified to map to a new output space, followed by fine-tuning the network parameters on the dataset of the target task.
Such techniques are especially useful when there is
a limited amount of training data available for the target task. 

Adversarial reprogramming shares the same objective as transfer learning with an additional constraint: the network architecture or parameters cannot be modified.
Instead, the adversary can only adapt the input and output interfaces of the network to perform the new adversarial task. This more constrained problem setting of adversarial reprogramming poses a security challenge to neural networks. An adversary can potentially re-purpose cloud-hosted machine learning (ML) models for new tasks thereby leading to theft of computational resources. Additionally, the attacker may reprogram models for tasks that violate the code of ethics of the service provider. 
For example, an adversary can repurpose a cloud-hosted ML API for solving captchas to create spam accounts.

Prior works on adversarial reprogramming~\cite{reprogramming,neekhara2019_adversarial,kloberdanz2020reprogramming,tsai2020transfer} have demonstrated success in repurposing Deep Neural Networks (DNNs) for new tasks using computationally inexpensive input and label transformation functions. 
One 
interesting finding of~\cite{reprogramming} is that neural networks can be reprogrammed even if the training data for the new task has no resemblance to the original data. The authors empirically demonstrate this by repurposing ImageNet~\cite{imagenet21k} classifiers on MNIST~\cite{mnist} digits with shuffled pixels showing that transfer learning does not fully explain the success of adversarial reprogramming. 
These results suggest that neural circuits hold properties that can be useful across multiple tasks which are not necessarily related. Hence neural network reprogramming not only poses a security threat, but also holds the promise of more reusable and efficient ML systems by enabling shared compute of the neural network backbone during inference time.

In 
existing work on adversarial reprogramming, the target adversarial task 
has
the same data domain as the original task.
Recent work has shown that network architectures based on the transformer model can achieve state-of-the-art results on language~\cite{NIPS2017_transformer}, audio~\cite{audiotransformer} and vision~\cite{dosovitskiy2021an} benchmarks suggesting that transformer networks serve as good inductive biases in various domains. 
Given this commonality between the neural architectures in different domains, an interesting question that arises is whether we can perform cross-modal adversarial reprogramming: For example, Can we repurpose a vision transformer model for a language task?

Cross-modal adversarial reprogramming increases the scope of target tasks for which a neural network can be repurposed. 
In this work, we develop techniques to adversarially reprogram image classification networks for discrete sequence classification tasks.
We propose a simple and computationally inexpensive adversarial program that embeds a sequence of discrete tokens into an image and propose techniques to train this adversarial program subject to a label remapping defined between the labels of the original and new task. 
We demonstrate that we can reprogram a number of image classification neural networks based on both Convolutional Neural Network (CNN)~\cite{lecun1998gradient} and Vision Transformer~\cite{dosovitskiy2021an} architectures to achieve competitive performance on a number of sequence classification benchmarks. 
Additionally, we show that it is possible to conceal the adversarial program as a perturbation in a real-world image thereby posing a stronger security threat. 
The technical contributions of this paper are summarized below:
\begin{itemize}
    \item We propose Cross-modal Adversarial Reprogramming, a novel approach to repurpose ML models originally trained for image classification to perform sequence classification tasks. To the best of our knowledge, this is the first work that expands adversarial reprogramming beyond the data domain of the original task.
    
    \item We demonstrate the feasibility of our method by re-purposing four image classification networks for six different sequence classification benchmarks covering sentiment, topic, and DNA sequence classification. Our results show that a computationally-inexpensive adversarial program can leverage the learned neural circuits of the victim model and outperform word-frequency based classifiers trained from scratch on several tasks studied in our work. 
    
    \item We demonstrate for the first time the threat imposed by adversarial reprogramming to the transformer model architecture by repurposing the Vision Transformer model for six different sequence classification tasks. The reprogrammed transformer model outperforms alternate architectures on five out of six tasks studied in our work. 

\end{itemize}



\section{Background and Related Work}



\begin{figure*}[h]
    \centering
    \includegraphics[width=1.0\linewidth]{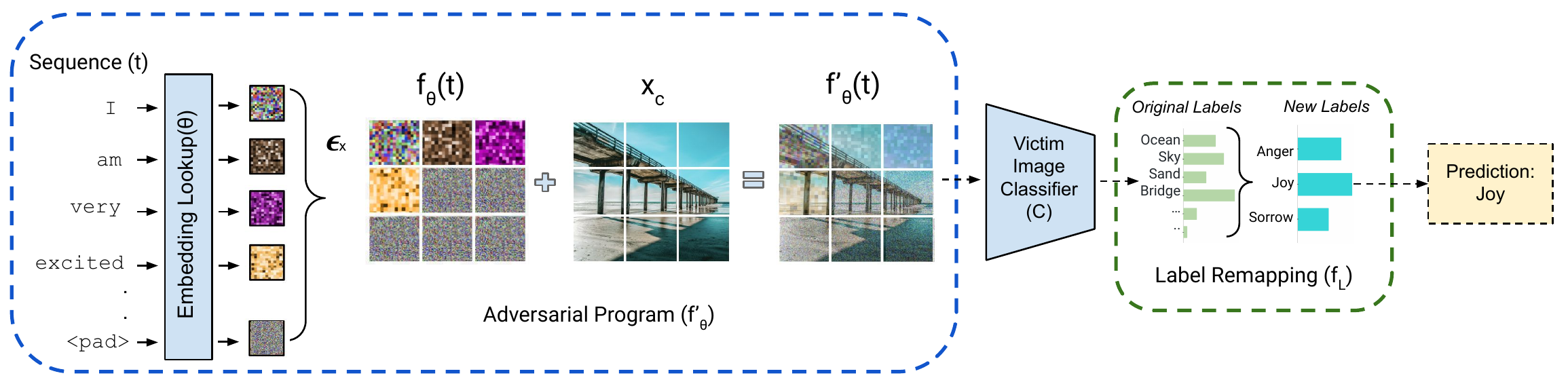}
    \caption{Schematic overview of our proposed cross-modal adversarial reprogramming method: The adversarial reprogramming function $f_\theta$ embeds a sequence of discrete tokens $t$ into an image. The image can also be concealed as an additive addition to some real-world image $x_c$ using the alternate reprogramming function $f'_\theta$. Finally, the victim model is queried with the generated image and the predicted label is mapped to the target label using the label remapping function $f_L$.
    }
    \label{fig:model_diagram}
\end{figure*}

\subsection{Adversarial Reprogramming}

Neural networks have been shown to be vulnerable to adversarial examples~\cite{harnessing,limitations,blackbox,universal,discrete1,tu2020physically,neekhara2019universal,waveguard,Hussain_2021_WACV,ACM_adv_deepfakes} which are slightly perturbed inputs that cause 
victim models to make a mistake.
Adversarial Reprogramming was introduced by~\cite{reprogramming} as a new form of adversarial threat that allows an adversary to repurpose neural networks to perform new tasks, which are different from the tasks they were originally trained for.
The proposed technique trains a single adversarial perturbation that can be added to all inputs in order to re-purpose the target model for an attacker's chosen task. The adversary achieves this by first defining a hard-coded one-to-one label remapping function that maps the output labels of the adversarial task to the label space of the classifier; and learning a corresponding adversarial reprogramming function that transforms an input from the input space of the new task to the input space of the classifier. The authors demonstrated the feasibility of their attack algorithm by reprogramming ImageNet classification models for classifying MNIST and CIFAR-10 data in a white-box setting, where the attacker has access to the victim model parameters. 

While the above attack does not require any changes to the victim model parameters or architecture, the adversarial program proposed~\cite{reprogramming} is only applicable to tasks where the input space of the the original and adversarial task is continuous. To understand the feasibility of attack in a discrete data domain,~\cite{neekhara2019_adversarial} proposed methods to repurpose \textit{text} classification neural networks for alternate tasks, which operate on sequences from a discrete input space. The attack algorithm used a context-based vocabulary remapping method that performs a computationally inexpensive input transformation to reprogram a victim classification model for a new set of sequences. This work was also the first in designing algorithms for training such an input transformation function in both white-box and black-box settings---where the adversary may or may not have access to the victim model’s architecture and parameters. They demonstrated the success of their proposed reprogramming functions by adversarially re-purposing various text-classification models including Long Short Term Memory
networks (LSTM)~\cite{THE_LSTM_PAPER}, bi-directional LSTMs~\cite{biLSTM} and CNNs~\cite{charCNN} for alternate text classification tasks.

Recent works~\cite{kloberdanz2020reprogramming,tsai2020transfer} have argued that reprogramming techniques can be viewed as an efficient training method and can be a superior alternative to transfer learning.
Particularly~\cite{tsai2020transfer} argue that one of the major limitations of current transfer learning techniques is the requirement of large amounts of target domain data, which is needed to fine-tune pre-trained neural networks. They demonstrated the advantage of instead using reprogramming techniques to repurpose existing ML models for alternate tasks, which can be done even when training data is scarce. 
The authors designed a black-box adversarial reprogramming method, that can be trained iteratively from input-output model responses, and demonstrated its success in repurposing ImageNet models for medical imaging tasks such as classification of autism spectrum disorders, melanoma detection, etc. 

All of these existing reprogramming techniques are only able to reprogram ML models when the data domain of the target adversarial task and the original task are the same. 
We address this limitation in our work by designing adversarial input transformation functions that allow image classification models to be reprogrammed for sequence classification tasks such as natural language and protein sequence classification. 

\subsection{Transformers and Image Classifiers}
While Convolutional Neural Networks (CNNs) have long achieved state-of-the-art performance on vision benchmarks, the recently proposed Vision Transformers (ViTs)~\cite{dosovitskiy2021an} have been shown to outperform CNNs on several image classification tasks. Transformers~\cite{NIPS2017_transformer} are known for achieving state-of-the-art performance in natural language processing (NLP).
In order to train transformers for image classification tasks, the authors~\cite{dosovitskiy2021an} divided an image
into patches and provide the sequence of linear embeddings of these patches as an input to a transformer. Image patches are treated the same way as tokens (words) in an NLP application and the model is trained on image classification in a supervised manner. The authors report that when ViTs are trained on large-scale image datasets, they are competitive and also outperform state-of-the-art models on multiple image recognition benchmarks. 

Since transformers can model both language and vision data in a similar manner, that is, as a sequence of embeddings, we are curious to 
investigate whether
a vision transformer can be reprogrammed for a text classification task. In the process, we find that CNN network architectures can also be reprogrammed to achieve competitive performance on discrete sequence classification tasks. In the next section, we discuss our cross-modal adversarial reprogramming approach.

\section{Methodology}

\subsection{Problem Definition}
Consider a victim image classifier $C$ trained for mapping images $x \in X$ to a label $\mathit{l_X} \in \mathit{L_X}$. That is, 
$$ C : x \mapsto \mathit{l_X} $$
An adversary wishes to repurpose this victim image classifier for an alternate text classification task $C'$ of mapping sequences $t \in T$ to a label $\mathit{l_T} \in \mathit{L_T}$. That is,
$$C' : t \mapsto l_T $$ 
To achieve this goal, the adversary needs to learn appropriate mapping functions between the input and output spaces of the original and the new task. We solve this by first defining a label remapping $\mathit{f_L}$ that maps label spaces of the two tasks:  $\mathit{f_L} : \mathit{l_X} \mapsto \mathit{l_T}$ ;  and then learning a corresponding adversarial program $\mathit{f_\theta}$ that maps a sequence $t \in T$ to an image $x \in X$  i.e.,~$\mathit{f_\theta}: t \mapsto x$ such that $\mathit{f_L}(C(f_\theta(t)))$ acts as the target classifier $C'$.

We assume a white-box adversarial reprogramming setting where the adversary has complete knowledge about architecture and model parameters of the victim image classifier. 
In the next few sections we describe the adversarial program $f_\theta$, the label remapping function and the training procedure to learn the adversarial program.

\subsection{Adversarial Program}
The goal of our adversarial program is to map a sequence of discrete tokens $t \in T$ to an image $x \in X$. 
Without loss of generalizability, we assume $X = [-1,1]^{\mathit{h\times w\times c}}$ to be the scaled input space of the image classifier $C$ where $\mathit{h,w}$ are the height and width of the input image and $c$ is the number of channels. The tokens in the sequence $t$ belong to some vocabulary list $V_T$. We can represent the sequence $t$ as $t = t_1,t_2,\ldots,t_N$ where $t_i$ is the vocabulary index of the $i_{\mathit{th}}$ token in sequence $t$ in the vocabulary list $V_T$.

When designing the adversarial program it is important to consider the computational cost of the reprogramming function $\mathit{f_\theta}$. This is because if a classification model that performs equally well can be trained from scratch for the classification task $C'$ and is computationally cheaper than the reprogramming function, it would defeat the purpose of adversarial reprogramming. 

Keeping the above in mind, we design a reprogramming function that looks up embeddings of the tokens $t_i$ and arranges them as contiguous patches of size $p \times p$ in an image that is fed as input to the classifier $C$. Mathematically, the reprogramming function $\mathit{f_\theta}$ is parameterized by a learnable embedding tensor $\theta_{|V_T| \times |p| \times |p| \times |c|}$ and performs the transformation $\mathit{f_\theta}: t \mapsto x$ as per Algorithm~\ref{alg:advprog}.

\begin{algorithm}
   \caption{Adversarial Program $\mathit{f_\theta}$}
   \label{alg:advprog}
\begin{algorithmic}
    \STATE {\bfseries Input:} Sequence $t = t_1,t_2,\ldots,t_N$
   \STATE {\bfseries Output:} Reprogrammed image $x_{h \times w \times c}$
   \STATE {\bfseries Parameters:} Embedding tensor $\theta_{|V_T| \times |p| \times |p| \times |c|}$
   \STATE $x \gets {0}_{h \times w \times c}$
   \FOR{ \textbf{each }$t_k$ in $t$}
    \STATE $i \gets \lfloor (k \times p)/h \rfloor$
    \STATE $j \gets  (k \times p) \bmod w $
	\STATE $x[i:i+p, j:j+p, :] \gets \tanh(\theta[t_k,:,:,:])$
   \ENDFOR
   \STATE {\bfseries return} $x$
\end{algorithmic}
\end{algorithm}

The patch size $p$ and image dimensions $h,w$ determine the maximum length of the sequence $t$ that can be encoded into the image. 
We pad all the input sequences $t$ all the way up to the maximum allowed sequence length with a padding token to fill up the reprogrammed image and clip any sequences longer than the maximum allowed length from the end. More details about the hyper-parameters can be found in our experiments section.

\textbf{Concealing the adversarial perturbation:}
Most 
past works on adversarial reprogramming have considered an unconstrained attack setting, where the reprogrammed image does not necessarily need to resemble a real-world image. However, as noted by~\cite{reprogramming}, it is possible to conceal the reprogrammed image in a real-world image by constraining the output of the reprogramming function. We can conceal the reprogrammed image as an additive perturbation to some real-world base image $x_c$ by defining an alternate reprogramming function $f'_\theta$ as follows:
\begin{equation}
  \mathit{f'_\theta}(t) = \mathit{Clip}_{[-1,1]}( x_{c} + \epsilon.\mathit{f_\theta}(t) )  
  \label{eq:boundedattack}
\end{equation}
Since the output of the original reprogramming function ${f_\theta}$ is bounded between $[-1, 1]$, we can control the $L_\infty$ norm of the added perturbation using the parameter $\epsilon \in [0, 1]$.

\textbf{Computational Complexity: } 
As depicted in Figure~\ref{fig:model_diagram}, during inference, the adversarial program only looks up embeddings of the tokens in the sequence $t$ and arranges them in an image tensor which can optionally be added onto a base image. 
Asymptotically, the time complexity of this adversarial program is linear in terms of the length of the sequence $t$. 
Since there are no matrix-vector multiplications involved in the adversarial program, it is computationally equivalent to just the embedding layer of a sequence-based neural classifier. Therefore the inference cost of the adversarial program is significantly less than that of a sequence-based neural classifier. Table 1 in our supplementary material compares the wall-clock inference time for a sequence of length $500$ for our adversarial program and various sequence-based neural classifiers used in our experiments.

\subsection{Label Remapping and Optimization Objective}
Past works~\cite{reprogramming,neekhara2019_adversarial,tsai2020transfer} on adversarial reprogramming assume that the number of labels in the target task are less than than the number of labels in the original task. In our work, we relax this constraint and propose label remapping functions for both of the following scenarios: 

\textbf{\textit{1. Target task has fewer labels than the original task:}}
Initial works on adversarial reprogramming defined a one-to-one mapping between the labels of the original and new task~\cite{reprogramming,neekhara2019_adversarial}. However, recent work~\cite{tsai2020transfer} found that mapping multiple source labels to one target label helps improve the performance over one-to-one mapping. Our preliminary experiments on cross-modal reprogramming 
confirm this
finding, however, we differ in the way the final score of a target label $l_t$ is aggregated---\cite{tsai2020transfer} obtained the final score for a target label as the mean of the scores of the mapped original labels. We found that aggregating the score by taking the maximum rather than the mean over the mapped original labels leads to faster training. Another advantage of using max reduction is that during inference, we can directly map the original predicted label to our target label without requiring access to probability scores of any other label. 

Consider a target task label $l_t$, mapped to a subset of labels $L_{S_t} \subset L_S$ of the original task under the many-to-one label remapping function $f_L$. We obtain the score for this target task label as the maximum of the scores of each label $l_i \in L_{S_t}$ by classifier $C$. 
That is,

\begin{equation}
   Z'_{l_t}(t) = \max_{l_i \in L_{S_t}} Z_{l_i}(f_\theta(t)),
   \label{eq:labelremapping}
\end{equation}
where $Z_k(x)$ and $Z'_k(t)$ represent the score (before softmax) assigned to some label $k$ by classifier $C$ and $C'$ respectively. 

To define the label remapping $f_L$, instead of randomly assigning $m$ source labels to a target label, we first obtain the model predictions on the base image $x_c$ (or a zero image in the case of an unbounded attack) and sort the labels by the obtained scores; We then assign the the highest scored source labels to each target label using a round-robin strategy until we have assigned $m$ source labels to each target label.

Note that while we need access to individual class scores during training (where we assume a white-box attack setting), during inference we can simply map the highest predicted label to the target label using the label remapping function $f_L$ without having to know the actual scores assigned to different labels. 

\textbf{\textit{2. Original task has fewer labels than the target task:}}
In this scenario, we map the probability distribution over the original labels to a distribution over target labels to class scores for the target label space using a learnable linear transformation. That is,
\begin{eqnarray}
& Z'(t) = \theta'_{|L_T|\times|L_X|} \cdot \mathit{softmax}(Z(f_\theta(t))).
\label{eq:multilabelmapping}
\end{eqnarray}

Here $Z'(t)$ is a vector representing class scores (logits) for the target label space. 
$\theta'_{|L_T|\times|L_X|}$ are the learnable parameters of the linear transformation that are optimized along with the parameters of the reprogramming function $f_\theta$. Note that unlike the previous scenario, in this setting, we assume that we have access to the probability scores of the original labels during both training and inference.

\textbf{Optimization Objective:} To train the parameters $\theta$ of our adversarial program, we use a cross-entropy loss between the target label and the model score predictions obtained as per Equation~\ref{eq:labelremapping} or Equation~\ref{eq:multilabelmapping}. We also incorporate an $L_2$ regularization loss for better generalization on the test set and to 
encourage more imperceptible perturbation in the case of our bounded attack. Therefore our final optimization objective is the following:
\begin{eqnarray}
& P_{l_t} = \mathit{softmax}(Z'(t))_{l_t} \nonumber \\
& E(\theta) = - \sum\limits_{t \in T} \log(P_{l_{t}}) + \lambda ||\theta||_2^2. \nonumber 
\label{eq:objective}
\end{eqnarray}
Here $\lambda$ is the regularization hyper-parameter and $P_{l_t}$ is the predicted class probability of the correct label $l_t$ for sequence $t$. We use mini-batch gradient descent using an Adam optimizer~\cite{adam} to solve the above optimization problem on the dataset of the target task.

\section{Experiments}

\subsection{Victim Image Classifiers}

To demonstrate cross-modal adversarial reprogramming, we perform experiments on four 
neural architectures trained on 
the
ImageNet dataset. 
We choose both CNNs and the recently proposed Vision Transformers (ViT)~\cite{dosovitskiy2021an} as our victim image classifiers. 
While CNNs have long achieved state-of-the-art performance on computer-vision benchmarks, the recently proposed ViTs have been shown to outperform CNNs on several image classification tasks. 
We choose the ViT-Base~\cite{dosovitskiy2021an},  ResNet-50~\cite{resnetpaper}, InceptionNet-V3~\cite{inceptionetpaper} and EfficientNet-B7~\cite{tan2019efficientnet} architectures. The details of these architectures are listed in Table~\ref{tab:victimmodels}. We perform experiments on both pre-trained and randomly initialized networks.

\setlength\tabcolsep{3pt}
\begin{table}[h]
\centering
\resizebox{\columnwidth}{!}{%
\begin{tabular}{@{}l|ccc|cc@{}}
  \multicolumn{4}{c}{} & 
  \multicolumn{2}{c}{\emph{Accuracy (\%)}}\\
\toprule
  Model & Abbr. & Type & \# Params & Top-1 & Top-5
\\ \midrule
ViT-Base & ViT  & Transformer & 86.9M &  84.2 & 97.2  \\
ResNet-50 & RN-50  & CNN & 25.6M & 79.0 & 94.4  \\
InceptionNet-V3 & IN-V3 & CNN & 23.8M & 77.5 & 93.5  \\
EfficientNet-B4 & EN-B4 & CNN & 19.3M & 83.0 & 96.3  \\
\bottomrule
\multicolumn{1}{c}{}
\end{tabular}
}
\caption{
Victim image classification networks used for adversarial reprogramming experiments. We include the number of parameters of each model and also the Top-1 and Top-5 test accuracy achieved on the ImageNet benchmark.}
\label{tab:victimmodels}
\end{table}

\setlength\tabcolsep{3pt}
\begin{table*}
\centering
\begin{tabular}{@{}l|ccrrcr|cc|cc@{}}
  \multicolumn{1}{c}{} & \multicolumn{6}{c}{\emph{Dataset Statistics}} & \multicolumn{4}{c}{\emph{Accuracy (\%)}}
  \\ \midrule
  &  &   &  &  &  & Avg & \multicolumn{2}{c|}{\emph{Neural Methods}} & 
  \multicolumn{2}{c}{\emph{TF-IDF}}  \\
  Dataset & Task Type &  \# Classes & \# Train & \# Test & Token & Length & Bi-LSTM & 1D-CNN & unigram & n-gram
   
\\ \midrule
Yelp & Sentiment & 2 & 560,000 & 38,000 & word & 135.6 & 95.94 & 95.18 & 92.50 & 92.93 \\
IMDB & Sentiment & 2 & 25,000 & 25,000 & word & 246.8 & 89.43 & 90.02 & 88.52 & 88.43 \\
\midrule
AG & Topic & 4 & 120,000 & 7,600 & word & 57.0 & 91.45 & 92.09 & 90.92 & 90.69 \\
DBPedia & Topic & 14 & 560,000 & 70,000 & word & 47.1 & 97.78 & 98.09 & 97.12 & 97.16 \\
\midrule
Splice & DNA & 3 & 2,700 & 490 & neucleobase & 60.0 & 93.26 & 83.87 & 51.42 & 72.24 \\
H3 & DNA & 2 & 13,468 & 1,497 & neucleobase & 500.0 & 86.84 & 85.43 & 75.68 & 78.89 \\
\bottomrule
\multicolumn{1}{c}{}
\end{tabular}
\caption{Statistics of the datasets used for our reprogramming tasks. We also include the test accuracy of both neural network based and TF-IDF based benchmark classifiers trained from scratch on the train set. }
\label{tab:datasetstats}
\end{table*}

\subsection{Datasets and Reprogramming Tasks}

In this work, we repurpose the aforementioned image classifiers for several discrete sequence classification tasks. 
We wish to analyze the performance of cross-modal adversarial reprogramming for different applications such as understanding language and analyzing sequential biomedical data.  Biomedical datasets e.g.~splice-junction detection in genes, often have fewer training samples than language based datasets and we aim to understand whether such limitations can adversely affect our proposed reprogramming technique. 

Sentiment analysis and topic classification are popular NLP tasks. However, analyzing the underlying semantics of the sequence is often not necessary for solving these tasks since word-frequency based statistics can serve as strong discriminatory features. In contrast, tasks like DNA-sequence classification requires analyzing the sequential semantics of the input and simple frequency analysis of the unigrams or n-grams does not achieve competitive performance on these tasks. To evaluate the effectiveness of adversarial reprogramming in both of these scenarios, we consider the following tasks and datasets in our experiments:

\subsubsection{Sentiment Classification}

1. Yelp Polarity Dataset \textbf{(Yelp)}~\cite{charCNN}: This is a dataset consisting of reviews from Yelp for the task of sentiment classification, categorized into binary classes of positive and negative sentiment.

\noindent 2. Large Movie Review Dataset \textbf{(IMDB)}~\cite{IMDB}: This is a dataset for binary sentiment classification of positive and negative sentiment from highly polar IMDB movie reviews. 

\subsubsection{Topic Classification}

1. AG's News Dataset \textbf{(AG)}~\cite{charCNN}: is a collection of more than 1 million news articles. News articles have been gathered from more than 2000 news sources and contains 4 classes: \textit{World, Sports, Business, Sci/Tech}.

\noindent 2. DBPedia Ontology Dataset \textbf{(DBPedia)}~\cite{charCNN}: consists of 14 non-overlapping categories from DBpedia 2014. The samples consist of the category and abstract of each Wikipedia article.

\setlength\tabcolsep{3pt}
\begin{table*}[h]
\centering
\begin{tabular}{@{}l|c|cccc|cccc|cccc@{}}
\multicolumn{2}{c}{} & \multicolumn{8}{c|}{\emph{Unbounded}} & \multicolumn{4}{c}{\emph{Bounded ($L_\infty=0.1$)}}\\
\midrule
  \multicolumn{2}{c|}{} & \multicolumn{4}{c|}{\emph{Pre-trained}} & 
  \multicolumn{4}{c|}{\emph{Randomly Initialized}} & \multicolumn{4}{c}{\emph{Pre-Trained}}
  \\ \midrule
  Task & $p$ & ViT & RN-50 & IN-V3 & EN-B4 & ViT & RN-50 & IN-V3 & EN-B4 & ViT & RN-50 & IN-V3 & EN-B4
\\ \midrule
Yelp  & 16 & 92.82	& \textbf{93.29} & 89.19 & 93.47 & 92.73 & 68.50	& 65.56 & 52.97	& 88.57 & 81.32 & 81.33 & 81.23\\
IMDB  & 16 & 86.76	& 85.60 & 80.67 & 87.26 & \textbf{88.38} & 81.08	& 52.87 & 50.26	& 82.07 & 72.28 & 71.22 & 81.42\\
\midrule
AG  & 16 & \textbf{91.59}	& 89.88 & 89.78 & 90.46 & 91.45 & 82.37	& 50.43 & 24.87	& 86.49 & 83.26 & 78.93 & 84.03\\
DBPedia  & 32 & \textbf{97.62}	& 96.31 & 95.70 & 96.77	 & 97.56  & 30.12	& 52.87 & 19.61 & 92.79 & 80.64 & 81.46 & 79.53\\
\midrule
Splice  & 48 & \textbf{95.31}	& 94.48 & 95.10 & 92.04 & 54.13 & 48.57	& 91.22 & 50.20	& 95.10 & 94.27 & 94.89  & 91.55\\
H3  & 16 & \textbf{82.57} & 78.16 & 80.29 & 80.16 & 77.02 & 73.00 & 64.20 & 51.17	& 76.62 & 72.01 & 75.55 & 75.42\\
\bottomrule
\multicolumn{1}{c}{}
\end{tabular}
\caption{Results (\% Accuracy on the test set) of adversarial reprogramming experiments targeting four image classification models for the six sequence classification tasks. In the unbounded attack setting, we target both pre-trained and randomly initialized image classifiers. In the bounded attack setting, the output of the reprogramming function is concealed as a perturbation (with $L_\infty$ norm of $0.1$) to a randomly selected ImageNet image shown in Figure~\ref{fig:reprogramming_example}.}
\label{tab:reprogrammingexps}
\end{table*}

\subsubsection{DNA Sequence Classification}

1. Splice-junction Gene Sequences (\textbf{Splice}): This dataset~\cite{splicedata,ucisplice} was curated for training ML models to detect splice junctions in DNA sequences. 
In DNA, there are two kinds of splice junction regions: Exon-Intron (EI) junction and Intron-Exon (IE) junction. This dataset contains sample DNA sequences of 60 base pair length categorized into 3 classes: ``EI" which contains exon-intron junction, ``IE" which
contains intron-exon junction, and ``N" which contain neither EI or IE regions. 

\noindent  2. Histone Protein Occupancy in DNA (\textbf{H3}): This dataset from~\cite{DNAdataCELL,dnaclassification} indicates whether certain DNA sequences wrap around H3 histone proteins. Each sample is a sequence with a length of 500 neucleobases. Positive samples contain DNA regions wrapping around histone proteins while negative samples do not contain such DNA regions. 

The statistics of these 
datasets are included in 
Table~\ref{tab:datasetstats}. To benchmark the performance that can be achieved on these tasks, we train various classifiers from scratch on the datasets for each task. We consider both neural network based classification models and frequency-based statistical models (such as TF-IDF) as our benchmarks. We use word-level tokens for sentiment and topic classification tasks and neucleobase level tokens for DNA sequence classification tasks.

The TF-IDF methods can work on either unigrams or n-grams for creating the feature vectors from the input data. 
For the n-gram model, 
we consider n-grams up to length 3 and choose the value of $n$ that achieves the highest classification accuracy on the hold-out set. We train a Stochastic Gradient Descent (SGD) classifier to classify the feature vector as one of the target classes. Additionally, we train DNN based text-classifiers: Bidirectional Long Short Term Memory networks (Bi-LSTM)~\cite{biLSTM,THE_LSTM_PAPER} and 1D CNN~\cite{kim2014convolutional} models from scratch on the above tasks. 
We use randomly initialized token embeddings for all classification models, which are trained along with the network parameters. For Bi-LSTMs, we combine the outputs of the first and last time step for prediction. For the Convolutional Neural Network we follow the same architecture as \cite{kim2014convolutional}. The hyper-parameter details of these classifiers and architecture have been included in Table 2 of the supplementary material.

We report the accuracies on the test set of the above mentioned classifiers in Table~\ref{tab:datasetstats}. We find that while both neural and frequency based TF-IDF methods work well on sentiment and topic classification tasks, neural networks significantly outperform frequency based methods on DNA sequence classification tasks. This is presumably because the latter require structural analysis of the sequence rather than relying on keywords.


\subsection{Experimental Details}
\textbf{Input image size and patch size:} The ViT-Base model utilized in our work is trained on images of size $384 \times 384$ and works on image patches of size $16 \times 16$. For all our experiments, we fix the input image size to be $384 \times 384$.
When we use a patch of size $16 \times 16$ for encoding a single token in our sequence, it allows for a maximum of $576$ tokens to be encoded into a single image. 
In our initial experiments we found that using larger patch sizes for smaller sequences leads to higher performance on the target task, since it encodes a sequence in a spatially larger area of the image. Therefore, we choose our patch size as the largest possible multiple of $16$ that can encode the longest sequence in our target task dataset. We list the patch size $p$ used for different tasks in Table~\ref{tab:reprogrammingexps}.

\textbf{Training hyper-parameters:}
We train each adversarial program on a single Titan 1080i GPU using a batch size of $4$. We set the learning rate as $0.001$ for the unbounded attacks and $0.001 \times \epsilon^{-1}$ for our bounded attacks (Equation~\ref{eq:boundedattack}). 
We set the $L_2$ regularization hyper-parameter $\lambda=1e-4$ for all our experiments and train the adversarial program for a maximum 100k mini-batch iterations in the unbouned attack setting and for 200k mini-batch iterations in the bounded attack setting. We map 10 original labels to each target label in the scenario when 
there are fewer labels for the target task than for the original task.
We point the readers to our codebase for precise implementation.\footnote{\url{https://github.com/paarthneekhara/multimodal_rerprogramming}}

\section{Results}
\subsection{Pre-trained vs untrained victim models}

Experimental results of our proposed cross-modal reprogramming method are reported in Table~\ref{tab:reprogrammingexps}. In these experiments, the original task has more labels than the target task so we use the label remapping function given by Equation~\ref{eq:labelremapping}. 
We first consider the unbounded attack setting, where the output of the adversarial program does not need to be concealed in a real-world image. For these experiments, we use the reprogramming function $f_\theta$ described in Algorithm~\ref{alg:advprog}.
We also note that the primary evaluation of past reprogramming works~\cite{reprogramming,neekhara2019_adversarial,tsai2020transfer} is done in an unbounded attack setting.

When attacking pre-trained image classifiers, we achieve competitive performance (as compared to benchmark classifiers trained from scratch, reported in Table~\ref{tab:datasetstats}) across several tasks for all victim image classification models. 
To assess the importance of pre-training the victim model on the original dataset, we also experiment with reprogramming untrained randomly initialized networks. 

Randomly initialized neural networks can potentially have rich structure which the reprogramming functions can exploit. 
Prior works~\cite{2018gaussian,lee2018deep} have shown that wide neural networks can behave as Gaussian processes, where training specific weights in the intermediate layers is not necessary to perform many different tasks. 
However, in our experiments, we find that for CNN-based image classifiers, reprogramming pre-trained neural networks performs significantly better than reprogramming randomly initialized networks for all tasks. 
This is consistent with the findings of prior reprogramming work~\cite{reprogramming} which reports that adversarial reprogramming in the image domain is more effective when it targets pre-trained CNNs. For the ViT model, we find that we are able to obtain competitive performance on sentiment and topic classification tasks when reprogramming either randomly initialized or pre-trained models. Particularly, we find that reprogramming untrained vision transformers provides the highest accuracy on the IMDB classification task. However, for DNA sequence classification tasks (Splice and H3) that require structural analysis of the sequence rather than token-frequency statistics, we find that reprogramming pre-trained vision transformer model performs significantly better than a randomly initialized transformer model. 

The ViT model outperforms other architectures on 5 out of 6 tasks in the unbounded attack setting. In particular, for the task of splice-junction detection in gene sequences, reprogramming a pre-trained ViT model outperforms both TF-IDF and neural classifiers trained from scratch. For sentiment analysis and topic classification tasks, which primarily require keyword detection, some reprogramming methods achieve competitive performance as the benchmark methods reported in Table~\ref{tab:datasetstats}. 

Additionally, to assess the importance of the victim classifier for solving the target task, we study the extent to which the task can be solved without the victim classifier and using only the adversarial reprogramming function with a linear classification head. We present the results and details of this experiment in Table 3 of our supplementary material.

\begin{figure}[h]
    \centering
    \includegraphics[width=1.0\columnwidth]{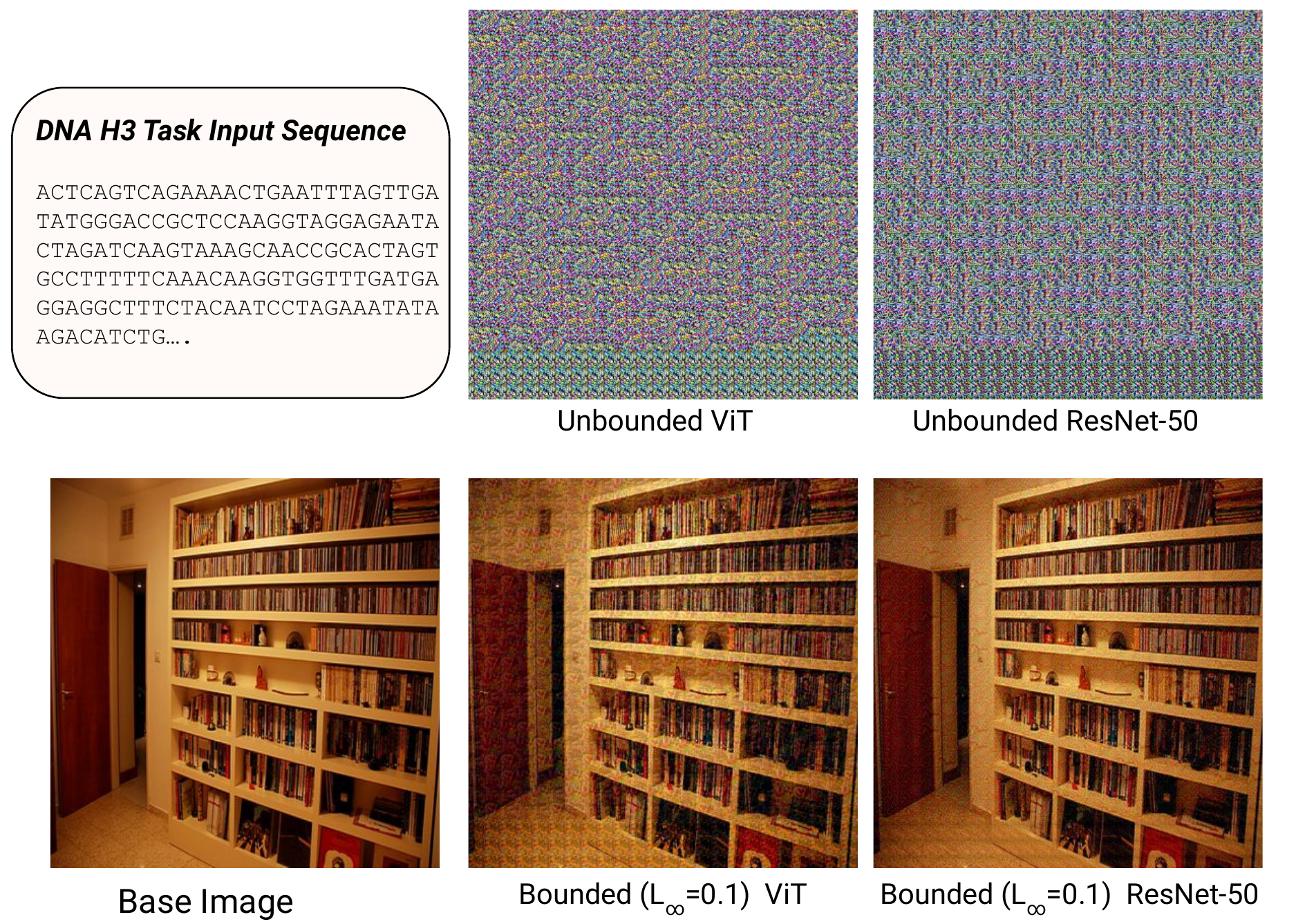}
    \caption{Example outputs of our adversarial reprogramming function in both unbounded (top) and bounded (bottom) attack settings while reprogramming two different pre-trained image classifiers for a DNA sequence classification task (H3).}
    \label{fig:reprogramming_example}
\end{figure}

\textbf{Concealing the adversarial perturbation: }
To conceal the output of the adversarial program in a real-world image, we follow the adversarial reprogramming function defined in Equation~\ref{eq:boundedattack}. We randomly select an image from the ImageNet dataset (shown in Figure~\ref{fig:reprogramming_example}) as the base image $x_c$
and train adversarial programs targeting different image classifiers for the same base image. We present the results at $L_\infty=0.1$ (on a 0 to 1 pixel value scale) distortion between the reprogrammed image and the base image $x_c$ on the right side of Table~\ref{tab:reprogrammingexps}. It can be seen that for some drop in performance, it is possible to perform adversarial reprogramming such that the input sequence is concealed in a real-world image. Figure 1 in our supplementary material shows the accuracy on three target tasks for different magnitudes of allowed perturbation, while reprogramming a pre-trained ViT model. 

\subsection{Target task has more labels than original task}

In a practical attack scenario, the adversary may only have access to a victim image classifier with fewer labels than the target task labels. 
To evaluate adversarial reprogramming in this scenario, we constrain the adversary's access to the class-probability scores of just $q$ 
labels of the ImageNet classifier. 
We choose the most frequent $q$ ImageNet labels as the original labels, that can be accessed by the adversary; 
and perform our experiments on two tasks from our datasets, which have the highest number of labels---AG News (4 labels) and DBPedia (14 labels).
We use the label remapping function given by Equation~\ref{eq:multilabelmapping}, and learn a linear transformation to map the predicted probability distribution over the $q$ original labels to the target task label scores. 

We demonstrate that we are able to perform adversarial reprogramming even in this more constrained setting. We achieve similar performance as compared to our many-to-one label remapping scenario reported in Table~\ref{tab:reprogrammingexps} when $q$ is close to the number of labels in the target task. This is because we learn an additional mapping function for the output interface, which can potentially lead to better optimization. However as a downside, this setting requires access to all $q$ class probability scores for predicting the adversarial label, while in the previous many-to-one label remapping scenario, we only need to know the highest-scored original label for mapping it to one of the adversarial labels. 

\begin{table}[h]
\centering
\begin{tabular}{@{}lcc|cccc@{}}
\multicolumn{3}{c}{} & 
  \multicolumn{4}{c}{\emph{Accuracy (\%)}}\\
\toprule
  Dataset & \# Labels & $q$ & ViT & RN-50 & IN-V3 & EN-B4
\\ \midrule
AG  & 4 & 3 & \textbf{89.42} & 87.18 & 86.66 & 89.18\\
DBPedia  & 14 & 3 & \textbf{96.34} & 83.16 & 84.17 & 92.95 \\
DBPedia  & 14 & 10 & \textbf{98.01} & 96.84 & 94.88 & 97.16\\
\bottomrule
\multicolumn{1}{c}{}
\end{tabular}
\caption{Results of adversarial reprogramming in the scenario when the target task has more labels than the original task. The access of the adversary is constrained to class-probabilities of $q$ labels of the original (ImageNet) task. This evaluation is done on pre-trained networks in an unbounded attack setting.}
\label{tab:multilabelexps}
\end{table}

\section{Conclusion}

We propose Cross-modal Adversarial Reprogramming, which for the first time demonstrates the possibility of repurposing pre-trained image classification models for sequence classification tasks. 
We demonstrate that computationally inexpensive adversarial programs can repurpose neural circuits to non-trivially solve tasks that require structural analysis of sequences.
Our results suggest the potential of training more flexible neural models that can be reprogrammed for tasks across different data modalities and data structures.
More importantly, this work reveals a broader security threat to public ML APIs that warrants the need for rethinking existing security primitives.

\section{Acknowledgements}
This work was supported by ARO under award number W911NF1910317, SRC under Task ID: 2899.001 and DoD UCR W911NF2020267 (MCA S-001364).


{\small
\bibliographystyle{ieee_fullname}
\bibliography{egbib2}
}
\newpage

\section*{}


\section*{Supplementary material}

# Cross-modal Adversarial Reprogramming - Supplementary Material

## 1. Wall-clock inference time of reprogramming function

In Table 1, we report the wall clock inference time for the adversarial program and the benchmark text classifiers studied in our work for a sequence of length 500. Both the Bi-LSTM and CNN model use 256 hidden units and an embedding size of 256. We use a single layer Bi-LSTM network and 1-D CNN with convolution filters of size 3, 4 and 5 based on the architecture proposed in (Kim, 2014). For the adversarial program the patch size is $16X16$ and the output image size is $384X384$. We average the inference time for 100 sequences for these evaluations. It can be seen that the adversarial program is significantly faster than both Bi-LSTM and 1D-CNN models for both CPU and GPU implementations in PyTorch. The CPU used for these evaluations is Intel Xeon CPU and the GPU is an Nvidia Titan 1080i.

| Model | CPU | GPU |
|---|---:|---:|
| Adversarial Program | 7.9 ms | 0.2 ms |
| Bi-LSTM | 161.5 ms | 13.9 ms |
| 1D CNN | 383.2 ms | 2.2 ms |

Table 1. Wall clock inference time (in miliseconds) for the adversarial program and the benchmark text classifiers studied in our work for a sequence of length 500.

## 2. Hyper-parameter details of benchmark classifiers

For training the benchmark neural-text classifiers, we use the Bi-LSTM and 1D-CNN model with a softmax classification head. For both of these models, the token embedding layer is randomly initialized and trained with the model parameters. We use Adam optimizer and perform mini-batch gradient descent using a batch size of 32 for both of these models for a maximum 200k mini-batch iterations. For the 1D-CNN models we use filters of size 3, 4 and 5 for the three convolutional layers. Other hyper-parameter details of these models are listed in Table 2.

| Model | Hidden Units | Emb. Size | # Layers | LR |
|---|---|---|---|---|
| Bi-LSTM | 256 | 256 | 1 | 1e-4 |
| 1D CNN | 256 | 256 | 3 | 1e-4 |

Table 2. Hyper-parameter details for the neural sequence classifiers used as benchmark classifiers in our work. LR: Learning Rate. Emb. Size: Embedding Size.

## 3. Perturbation amount vs Accuracy in Bounded attacks

Figure 1 shows the performance of cross-modal adversarial reprogramming at different magnitudes of allowed perturbation in the bounded attack setting, while attacking the ViT model. We use the same base image $x_c$ as used in all of our bounded attack experiments.

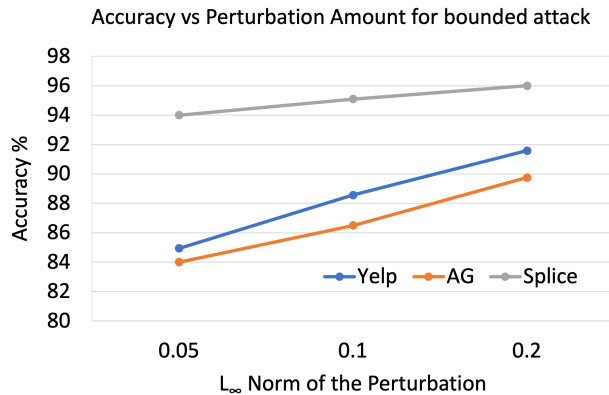

Figure 1. Accuracy vs $L_\infty$ norm of the perturbation while reprogramming the ViT model for three target tasks covering emotion, topic and DNA sequence classification.

## 4. Assessing the importance of Victim Model

To assess the importance of the victim model for solving the target task, we perform an experiment to understand the extent to which the target task can be solved by using only the adversarial reprogramming function with a linear classification head on top. To perform this experiment, we take the mean of all token embeddings in a sequence and pass it as input to a linear layer that predicts the class scores for the target task. Not surprisingly, this classifier works well for



\end{document}